\documentclass[final]{cvpr}

\usepackage{times}
\usepackage{epsfig}
\usepackage{graphicx}
\usepackage{amsmath}
\usepackage{amssymb}
\usepackage{booktabs}
\usepackage{multirow}
\usepackage{color}
\usepackage{array}
\usepackage[ruled]{algorithm2e}
\usepackage{url}

\def\tdr#1{${\rm TDR_{{#1}\%}}$}

\usepackage[pagebackref=true,breaklinks=true,colorlinks,bookmarks=false]{hyperref}

\def\cvprPaperID{409} 
\def\confYear{CVPR 2021}

\begin{document}

\title{Representative Forgery Mining for Fake Face Detection}
\author{Chengrui Wang, Weihong Deng*\\
Beijing University of Posts and Telecommunications\\
{\tt\small \{crwang, whdeng\}@bupt.edu.cn}
}

\maketitle

\begin{abstract}
    Although vanilla Convolutional Neural Network (CNN) based detectors can achieve satisfactory performance on fake face detection, we observe that the detectors tend to seek forgeries on a limited region of face, which reveals that the detectors is short of understanding of forgery.
    Therefore, we propose an attention-based data augmentation framework to guide detector refine and enlarge its attention.
    Specifically, our method tracks and occludes the Top-N sensitive facial regions, encouraging the detector to mine deeper into the regions ignored before for more representative forgery.
    Especially, our method is simple-to-use and can be easily integrated with various CNN models.
    Extensive experiments show that the detector trained with our method is capable to separately point out the representative forgery of fake faces generated by different manipulation techniques, and our method enables a vanilla CNN-based detector to achieve state-of-the-art performance without structure modification.
    Our code is available at \url{https://github.com/crywang/RFM}.
\end{abstract}
\section{Introduction}

The rapid development of face manipulation technology makes the manufacture of fake face more accessible than before, which further accelerates the spread of fake facial images on social media~\cite{deepfakes,faceapp,suwajanakorn2017synthesizing,thies2016face2face}.
Meanwhile, advanced techniques make it extremely difficult for human to distinguish between real and fake face~\cite{thispersondoesnotexist,dang2020detection}, raising constant concerns about the credibility of digital content~\cite{roessler2018faceforensics,roessler2019faceforensicspp,shu2017fake}.
To mitigate the toll that manipulation technology takes on society, Convolutional Neural Network (CNN) is widely used to construct detector for fake face detection~\cite{tolosana2020deepfakes}.

Unfortunately, although vanilla CNN-based fake face detector can achieve satisfactory detection performance~\cite{rossler2019faceforensics,tolosana2020deepfakes}, it may have a different understanding of forgery against that of humans.
Concretely, the experiment in Section~\ref{sec:visualization} shows that vanilla CNN-based detector tends to check forgeries from a limited region of face, while humans usually find representative forgery over the entire face.
For example, the representative forgeries of fake face generated by Deepfakes~\cite{deepfakes} and Face2Face~\cite{thies2016face2face} usually appear on the facial boundary, while the representative forgeries of fake face generated by StyleGAN~\cite{karras2019style} and PGGAN~\cite{karras2018progressive} are located in the entire face flexibly (shown in Figure~\ref{fig:exampleimg}).

\begin{figure}[t]
    \begin{center}
        \includegraphics[width=0.98\linewidth]
        {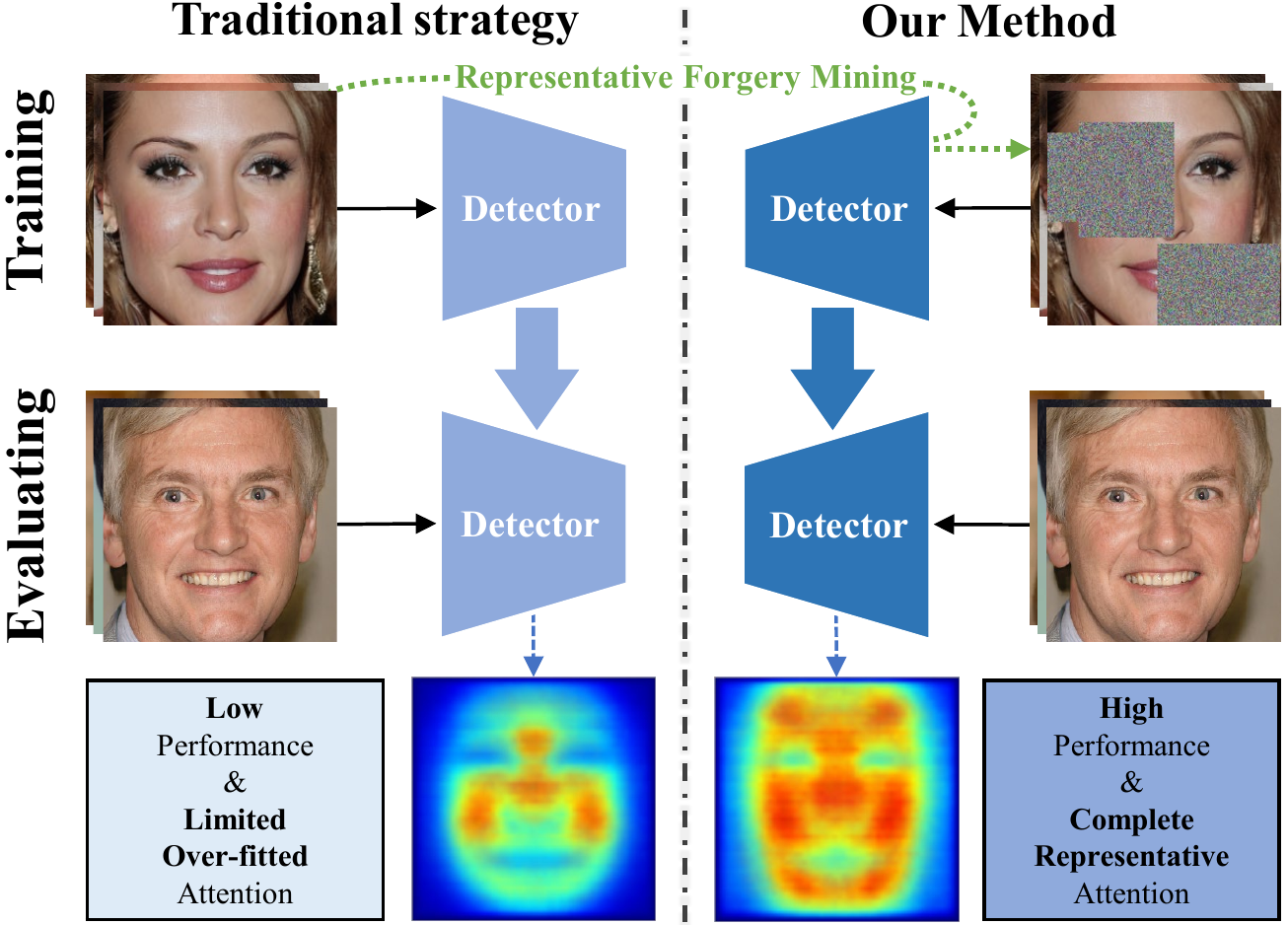}
    \end{center}
    \vspace{-0.1cm}
    \caption{
        Comparison of detectors trained with traditional strategy and with our proposed RFM.
        Through refining training data, our method provides meaningful guidance to force a vanilla detector to allocate its attention to a larger and representative facial region, which greatly improves detection performance.
        Meanwhile, the region of interest raises a remarkable correlation to the corresponding manipulation technique.
    }
    \label{fig:simpleprocess}
    \vspace{-0.5cm}
\end{figure}

For better detection, detectors should allocate more attention to the forgeries which can significantly represent the corresponding manipulation technique, rather than overfitting the forgeries which are mainly useful in minimizing the bi-classification loss function on training set.
Recent remarkable breakthroughs~\cite{chai2020makes,chen2020manipulated,dang2020detection,huang2020fakelocator,li2020face,zhang2019detecting,wang2020cnn,yu2019attributing,liu2020global} address this problem to some extent, which generally follow three directions:
a) Through extracting the digital fingerprints produced by the defect of manipulation technique, some works~\cite{liu2020global,wang2020cnn,yu2019attributing, zhang2019detecting} achieve advanced generalization performance on CNN-generated facial images;
b) Some works~\cite{chai2020makes,chen2020manipulated} divide face into multiple patches and detect them independently, which compulsively optimizes the receptive field of detectors on fake face.
c) With well-designed training dataset, some works~\cite{dang2020detection,huang2020fakelocator,li2020face} leverage the difference between real and fake faces of the same source to guide detector learn the forgeries on fake faces, which can further achieve forgery visualization.

In this paper, we propose an attention-based data augmentation method Representative Forgery Mining (RFM) to address the limited-attention problem by refining training data during training process.
A brief comparison between the traditional strategy and our method is sketched in Figure~\ref{fig:simpleprocess}.
Concretely, our method consists of two steps including 1) using the gradient of detector to generate image-level Forgery Attention Map (FAM), which can precisely locate the sensitive facial region, and 2) utilizing Suspicious Forgeries Erasing (SFE) to intentionally occlude the Top-N sensitive regions of face, allowing detector to explore representative forgery from the previously ignored facial region.
Specially, our method can be easily integrated with various CNN models without extra structure modification and sophisticated training set.

Through decoupling detector's attention from the over-sensitive facial region, our method achieves competitive detection performance with state of the art, and significantly maintains the detection performance on fake faces which only contain few technical forgeries.
Moreover, the region of interest visualized by average FAM shows that our method contributes to mining the representative forgery of different manipulation techniques.
The main contributions of this work are as follows:

\begin{figure}[t]
    \begin{center}
        \includegraphics[width=0.99\linewidth]
        {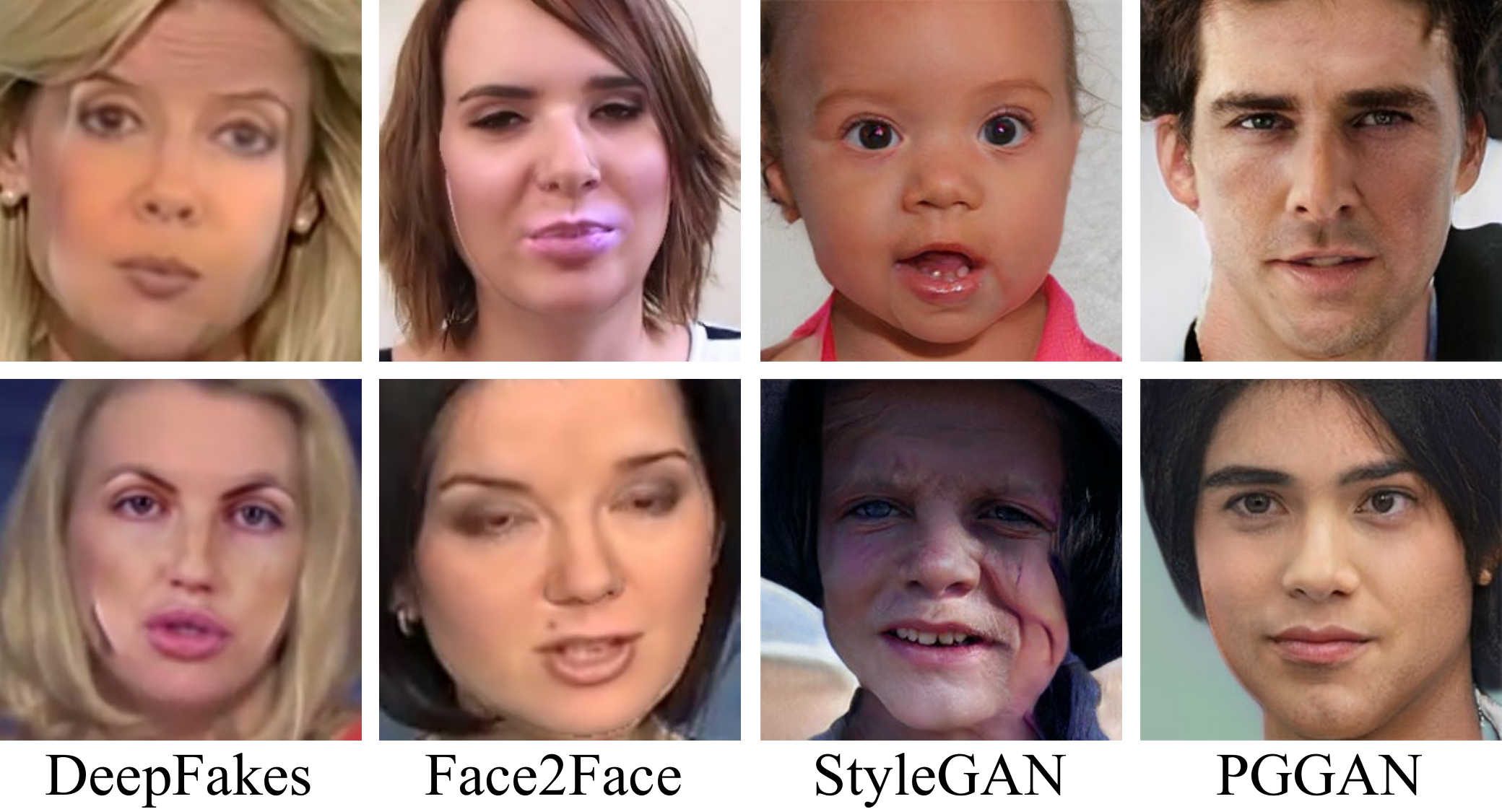}
    \end{center}
    \vspace{-0.3cm}
    \caption{
        Examples of fake faces generated by different manipulation techniques.
        The faces generated by Deepfakes and Face2Face have forgeries mainly on the facial boundary, while the forgeries of faces generated by StyleGAN and PGGAN could appear anywhere on the face.
    }
    \label{fig:exampleimg}
    \vspace{-0.4cm}
\end{figure}


– We propose a tracer method called FAM to precisely locate the facial region to which detector is sensitive, and further use it as the guidance for data augmentation.

– We propose an attention-based data augmentation method called SFE to help detector allocate more attention to representative forgery under the guidance of FAM.

– We finally provide a framework called RFM, which visualizes representative forgery without well-designed supervision and enables a vanilla CNN-based detector to achieve SOTA performance on DFFD and Celeb-DF.




\section{Related Work}
\textbf{Face Manipulation Techniques.}
According to technical procedure, well-known face manipulation techniques~\cite{deepfakes, faceapp, faceswap, choi2018stargan, karras2018progressive, karras2019style, karras2020analyzing,suwajanakorn2017synthesizing, thies2016face2face} can be roughly divided into one-stage and two-stage technique:

The main procedure of \textbf{two-stage technique}~\cite{deepfakes, faceswap, li2020advancing, suwajanakorn2017synthesizing, thies2016face2face} can be briefly described as: 1) generating target face or extracting face from target individual, 2) blending target identity into source face by utilizing mask, graphics-based technique, etc.
In practice, two-stage techniques are widely used for identity swap and expression manipulation.
Concretely, Thies \etal~\cite{thies2016face2face} propose a method to transfer facial expressions from target to source face while maintaining the identity of source person.
``Synthesizing Obama''~\cite{suwajanakorn2017synthesizing} composites synthesized mouth texture with proper 3D pose to help source face match the mouth in target video.
\textit{FaceSwap}~\cite{faceswap} can swap the face of a person seen by camera with the face in the provided image.
\textit{Deepfake}~\cite{deepfakes} is the symbol of CNN-based face identity swap, which uses an autoencoder to swap the identity of face.
Li \etal~\cite{li2020celeb} generate a large-scale fake face dataset by using improved synthesis process, solving the low resolution, color mismatch, inaccurate face masks and temporal flickering problems.
Li \etal~\cite{li2020advancing} propose a two-stage algorithm, achieving high fidelity and occlusion aware face swapping.

Most of \textbf{one-stage techniques}~\cite{faceapp, choi2018stargan, karras2018progressive, karras2019style, karras2020analyzing} are implemented based on GANs, which can achieve entire face synthesis or expression and attributes manipulation without constructing complex physical models.
\textit{FaceApp}~\cite{faceapp} is a consumer-level mobile application, providing multiple filters to selectively modify facial attributes.
Choi \etal~\cite{choi2018stargan} achieve facial attribute transfer and facial expression synthesis for multiple domains by using only a single model.
ProGAN~\cite{karras2018progressive} is a popular GAN structure that can synthesize high-resolution facial images by progressively growing both generator and discriminator.
StyleGAN~\cite{karras2019style} achieves the control of synthesis through a new structure, which can automatically separate the high-level attributes and stochastic variation in generated images.
Karras \etal~\cite{karras2020analyzing} further improve the perception quality of synthesized images by redesigning the structure of StyleGAN.

\begin{figure*}[t]
    \begin{center}
        \includegraphics[width=0.99 \linewidth]
        {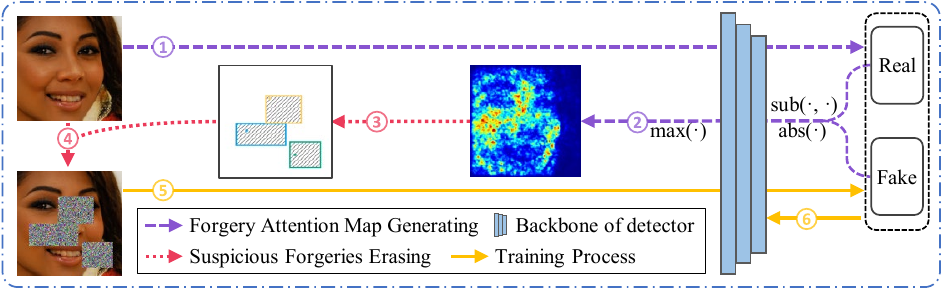}
    \end{center}
    \vspace{-0.2cm}
    \caption{
        The procedure of RFM, which can be divided into three parts.
        Firstly (in steps 1, 2), we generate FAM for each original image of a single mini-batch.
        Then (in steps 3 and 4), we utilize SFE to erase the original images under the guidance of FAMs generated before.
        Finally (in steps 5 and 6), we use the erased images as inputs to train detector.
        Specially, in contrast to offline pre-processing, RFM refines training data dynamically during training.
    }
    \label{fig:process}
    \vspace{-0.4cm}
\end{figure*}

\textbf{Fake Face Detection.}
Recent studies~\cite{chai2020makes,chen2020manipulated,dang2020detection,huang2020fakelocator,li2020face,liu2020global,nataraj2019detecting,wang2020cnn,yu2019attributing,zhang2019detecting} propose a variety of methods for fake face detection.
Dang \etal~\cite{dang2020detection} assemble an attention-based layer into detector to locate forgery region and improve detection performance.
Huang \etal~\cite{huang2020fakelocator} locate forgery region by using a modified semantic segmentation network.
Chen \etal~\cite{chen2020manipulated} propose a detector which combines both spatial domain and frequency domain as inputs for detection.
Chai \etal~\cite{chai2020makes} modify the structure of Xception~\cite{chollet2017xception}, regarding each receptive field as a patch and detecting them independently.
Li \etal~\cite{li2020face} achieve high generalization detection performance without using fake images generated by any existed manipulation methods.
Specially, a series of methods~\cite{liu2020global,nataraj2019detecting,wang2020cnn,yu2019attributing,zhang2019detecting} are proposed to detect GAN-generated fake face by leveraging the detectable digital fingerprints produced by generators.

\textbf{Data Augmentation.}
Data augmentation is a useful approach in addressing the underfitting problem caused by insufficient data and preventing network from overfitting during training process~\cite{devries2017improved,krizhevsky2012imagenet,simonyan2014very,srivastava2014dropout,wei2017object,zhang2018mixup,zhong2020random}.
The most commonly used random cropping~\cite{krizhevsky2012imagenet} and random flip~\cite{simonyan2014very} extracts a random patch from the original image and randomly flips the original image, respectively.
Dropout~\cite{srivastava2014dropout} can also be regarded as a data augmentation method, which randomly selects some hidden neurons and sets their outputs to zero.
Mixup~\cite{zhang2018mixup} creates new images to expand dataset by calculating the weighted average of two different images.
Cutout~\cite{devries2017improved} applies a fixed-size zero-mask to a random location in image, while Random Erasing~\cite{zhong2020random} randomly selects a rectangle region in image and masks the region with random integers.
Adversarial Erasing~\cite{wei2017object} selectively masks image based on the guidance of Class Activation Mapping (CAM)~\cite{zhou2016learning}.

\section{Proposed Method}
In this section, we propose an attention-based data augmentation method called \textit{Representative Forgery Mining} (RFM).
As shown in Figure~\ref{fig:process}, our method is composed of two components.
1) \textit{Forgery Attention Map} (FAM) is the foundation of RFM, which can reveal the sensitivity of detector on each facial region.
2) Based on FAM, \textit{Suspicious Forgeries Erasing} (SFE) is applied to augment the original image for detector training.
During training, each iteration with RFM only needs to propagate forward and backward twice.
In the rest of this section, we explain the main difference between RFM and well-known erasing methods~\cite{wei2017object, zhong2020random}.
Same as common settings, we take fake face detection as a binary classification problem.

\subsection{Forgery Attention Map}\label{sec:msm}

In order to achieve guided erasing and forgeries visualization, FAM is proposed to precisely locate the region to which detector is sensitive.
Concretely, the most sensitive region is defined as the region where perturbation has the most critical impact on detection result.
In forward propagation, detector receives facial image $I$ as input and outputs two logits $O_{real}$ and $O_{fake}$ to measure whether $I$ is real or not.
Because any perturbation would affect both two logits, the detection result should be determined by the relative magnitude of the two logits.
By utilizing the $\nabla_I{O_{real}}$ and $\nabla_I{O_{fake}}$ to separately represent how perturbation in $I$ impacts on the logits outputs, the maximum absolute difference between $\nabla_I{O_{real}}$ and $\nabla_I{O_{fake}}$ is regarded as FAM to simply represent the impact of perturbation on detection result.
In other words, each value in FAM precisely indicates the sensitivity of detector to the corresponding pixel in image.
Formally, FAM $Map$ can be formulated as
\begin{equation}
    \begin{split}
        \textit{Map}_{I} = max\big(&abs(\nabla_I{O_{fake}}-\nabla_I{O_{real}})\big)\\
        = max\Big(&\nabla_I\big(abs({O_{fake}}-{O_{real}})\big)\Big)\ ,
    \end{split}
\end{equation}
where the function $max(\cdot)$ calculates the maximum value along channel axis and the function $abs(\cdot)$ obtains the absolute value of each pixel.

The difference between FAM and well-known Class Activation Mapping~\cite{chattopadhay2018grad,selvaraju2017grad,zhou2016learning} can be demonstrated from two aspects.
On the one hand, FAM locates the region where detector is sensitive, while Class Activation Mapping highlights the region which detector used for decision-making.
On the other hand, FAM generates map at image level, while Class Activation Mapping calculates map based on the last convolutional layer of network.

\begin{figure}[t]
    \setlength{\abovecaptionskip}{0pt}
    \setlength{\belowcaptionskip}{0pt}
    \begin{center}
        \includegraphics[width=0.99\linewidth]
        {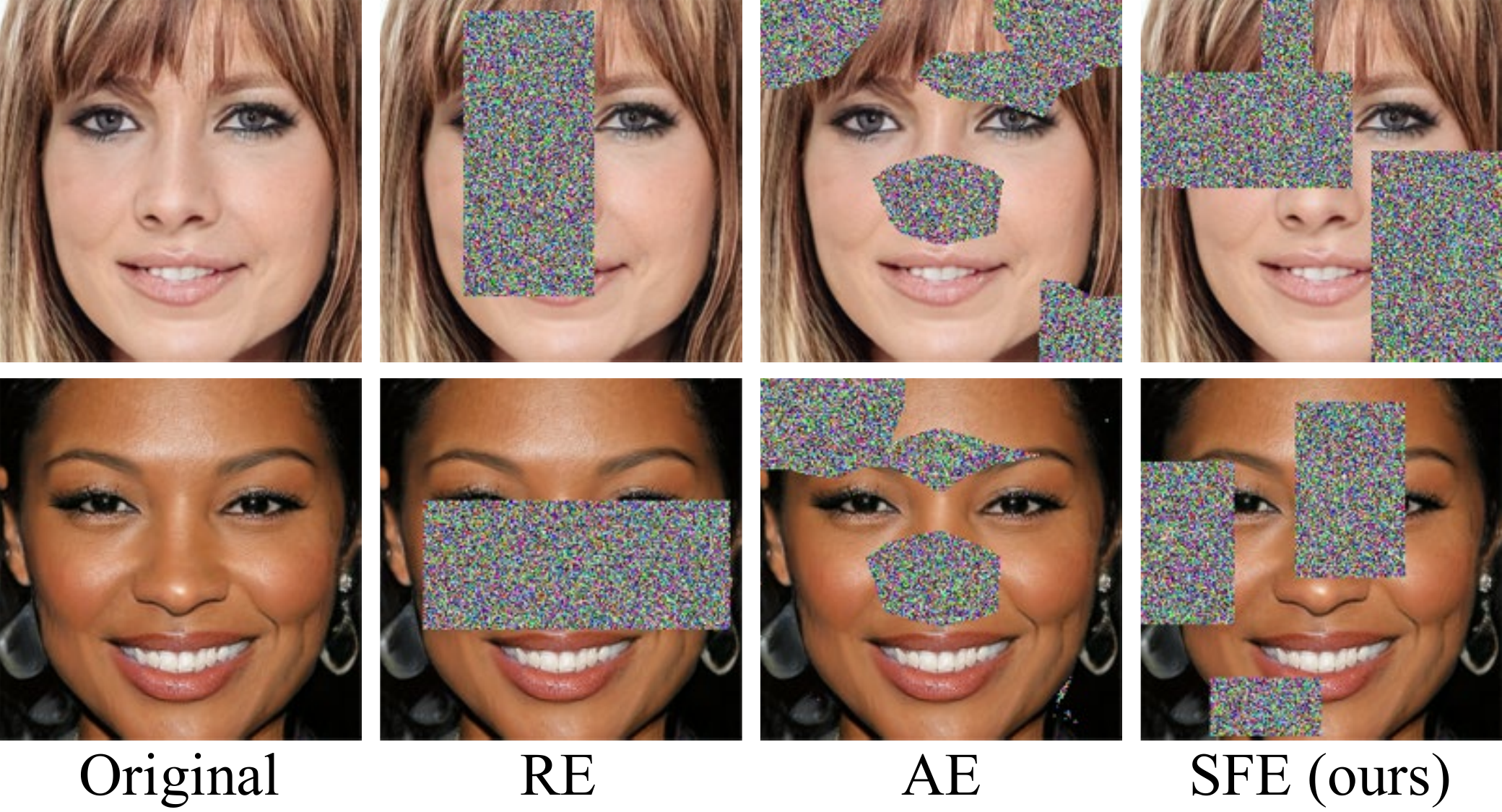}
    \end{center}
    \vspace{-0.2cm}
    \caption{
        Examples of faces processed by RE, AE, and SFE.
        Obviously, our SFE is more flexible than other methods.
    }
    \label{fig:RFM}
    \vspace{-0.3cm}
\end{figure}

\subsection{Suspicious Forgeries Erasing}\label{sec:SFE}
Through occluding the Top-N sensitive facial regions calculated by FAM, our proposed erasing method SFE realizes dynamic refinement.
In detail, we firstly generate FAM for each image in mini-batch.
The sizes of both FAM and input image can be assumed as $ H \times W $.
Then, for each image, we sort coordinates in descending order according to the values in the corresponding FAM generated before.
Next, each pixel is treated as an anchor according to the order calculated above.
For each anchor, we use random integers to form a rectangle block whose size is smaller than $H_e \times W_e$ ($H_e \leq H, W_e \leq W$) to occlude the anchor if it has not been occluded before.
We repeat the occlusion process until each image has been occluded by $N$ blocks.
The detail procedure of SFE is summarized in Algorithm~\ref{alg:SFE}.

\begin{algorithm}[t]
    \small
    \caption{Suspicious Forgeries Erasing}
    \label{alg:SFE}
    \LinesNumbered
    \KwIn{\quad Input facial image $I$\;
        \qquad \quad \quad Image size $H$ and $W$\;
        \qquad \quad \quad Forgery Attention map $Map$\;
        \qquad \quad \quad Erasing Block count $N$\;
        \qquad \quad \quad Erasing probability $p$\;
        \qquad \quad \quad Max erase size $H_{max}$ and $W_{max}$}
    \KwOut{\ Erased image $I^{*}$.}

    \If{Rand(0,1) $\leq$ p}{
    $cnt = 0$\;
    \While{cnt $<$ N}{
    $[i, j] = $ coordinate of the $ind^{th}$ largest value in $Map$\;
    \If{I[i, j] has not been occluded}{
    $H_{t} = Rand(1, H_{max})$\;
    $W_{l} = Rand(1, W_{max})$\;
    $H_{b} = H_{max} - H_{t}$\;
    $W_{r} = W_{max} - W_{l}$\;
    Fill $I[i-H_{t}:i+H_{b}, j-W_{l}:j+W_{r}]$ with a block composed of random integers\;
    $cnt = cnt+1$\;
    }
    }
    }
    $I^{*} \leftarrow I$\;
    \Return{$I^{*}$}\;
\end{algorithm}

\subsection{Comparison with well-known erasing methods.}
To demonstrate why SFE is recommended for fake face detection, we take an analysis on the main difference between SFE and well-known erasing methods such as Random Erasing (RE)~\cite{zhong2020random} and Adversarial Erasing (AE)~\cite{wei2017object}.

Random Erasing partially occludes image at a random position with a single random-sized rectangle mask, making network robust to single occlusion.
However, in term of fake face detection, the forgeries on which detector focuses may be far away from each other, causing it difficult for RE to erase all the forgeries while retaining as much effective facial information as possible.
Meanwhile, due to the lack of effective guidance, RE can not selectively erase the forgeries which would lead to overfitting.
Additionally, the inherent algorithm defect of RE makes RE more inclined to erase the central region of image.

Adversarial Erasing is a guided method that can progressively erase the discriminative object region.
However, the Class Activation Mapping~\cite{zhou2016learning} which AE utilized to locate the erasing region is calculated on the last convolutional layer of network, which may cause the occlusion position to be different from the region that should be occluded.
Moreover, the mask generating method which AE used is so fine-grained that may increase the risk of overfitting the shape and location of the mask.

In comparison with these methods, SFE can
1) precisely occlude sensitive facial regions under the guidance of FAM,
2) utilize multiple blocks to flexibly erase forgeries of different locations and preserve as much facial region as possible,
3) never leak extra information to detector and prevent detector from overfitting to the shape or location of erasing block.
The example facial images produced by different erasing methods are illustrated in Figure \ref{fig:RFM}.

\section{Experiments}
\subsection{Dataset}
We evaluate our method by performing experiments on two well-known datasets: DFFD~\cite{dang2020detection} and Celeb-DF~\cite{li2020celeb}.

\textbf{DFFD}~\cite{dang2020detection} contains 58,703 real facial images and 240,336 fake facial images.
The manipulation techniques in DFFD are various in category, including face identity swap, face expression and attributes manipulation, and entire face synthesis.
Moreover, both one-stage and two-stage manipulation techniques are used to generate fake faces in DFFD.

Specially, due to the inaccessibility of DFL~\cite{deepfacelab}, the DFFD we collected does not contain DFL database.
According to the number of manipulation technical stages, we divided the fake faces in DFFD into Group A and Group B.
In detail, \textbf{Group A} contains fake faces generated by two-stage techniques such as FaceSwap~\cite{faceswap}, Deepfakes~\cite{deepfakes} and Face2Face~ \cite{thies2016face2face}, while the manipulation techniques of face images in \textbf{Group B} is composed of the one-stage techniques such as FaceAPP\cite{faceapp}, StarGAN~\cite{choi2018stargan}, PGGAN~\cite{karras2018progressive} and StyleGAN~\cite{karras2019style}.
It must be pointed out that the images in Group A are collected from \textbf{FaceForensics++}~\cite{roessler2019faceforensicspp}.

\textbf{Celeb-DF}~\cite{li2020celeb} is a symbol of the second generation manipulation technology, which generates fake face through a improved two-stage technique.
Celeb-DF contains 590 real videos collected from YouTube video clips of 59 celebrities and 5,639 high-quality fake videos of celebrities generated using improved synthesis process.
The fake faces in Celeb-DF are more difficult to distinguish than the fake faces in the previous datasets of the same category.
For fake face detection, we extract facial images from the key frames of videos in Celeb-DF.

\subsection{Experiment Settings}
We firstly resize the aligned facial images into a fixed size of $256 \times 256$.
Then, we apply random and center cropping into training and testing process to resize the images to $224 \times 224$, respectively.
Moreover, we flip each image horizontally with a probability of 50\% during training.

We adopt Xception~\cite{chollet2017xception} as the backbone of detector.
All the detectors are trained by using Adam~\cite{kingma2015adam} optimizer with fixed learning rate of 0.0002.
Following~\cite{dang2020detection}, the size of mini-batch is set to 16, and each mini-batch consists of 8 real and 8 fake facial images.
On the basis of cross entropy loss function, we extra utilize the loss term in~\cite{ishida2020we} with $b = 0.04$ to stabilize training.
To fairly compare performance of the detectors trained with and without our method, all the detectors are trained from a same weight initialization.
The hyper-parameters $N$, $p$, $H_{max}$ and $W_{max}$ are implicitly set as 3, 1.0, 120 and 120, respectively.

We report the detection performance by using the evaluation metrics such as Area Under Curve (AUC) of ROC, True Detect Rate (TDR) at False Detect Rate (FDR) of 0.01\% (denoted as \tdr{0.01}), and TDR at FDR of 0.1\% (denoted as \tdr{0.1}).

\subsection{Ablation Study}
In this section, we perform a number of ablation studies to better understand the contribution of each component and hyper-parameter in RFM.

\textbf{Effect of Forgery Attention Map and Multiple Erasing Blocks.}
We conduct experiments on DFFD to investigate how Forgery Attention Map (FAM) and Multiple Erasing Blocks (MEB) boost detection performance.
The functions of FAM and MEB are independent in RFM, where FAM plays the role of guidance and MEB emphasizes erasing with multiple blocks.
The results are shown in Table~\ref{tab:ablation}, where ``FAM\&MEB'' is the original setting, ``w/ MEB'' denotes placing the anchors of SFE randomly, ``w/ FAM'' denotes only occluding the Top-1 sensitive region under the guidance of FAM, and ``w/o\ MEB$|$FAM'' denotes using a single erasing block to occlude a random region of face.

Although the ordinary erasing (``w/o\ MEB$|$FAM'') can improve detection performance, we observe that MEB and FAM further improve \tdr{0.01} by 0.51\% and 2.21\% on DFFD, respectively, demonstrating that either MEB or FAM has contribution to our algorithm and FAM is more effective than MEB.
Compared with baseline detector, the detector trained with combining both MEB and FAM leads to significant improvements of 8.33\% \tdr{0.01} and 3.88\% \tdr{0.1} on DFFD, which also outperforms all the other settings by a large margin.

\begin{table}[t]
    \small
    \begin{center}
        \begin{tabular}{l|c|c|c}
            \toprule
            Method                          & AUC            & \tdr{0.1}      & \tdr{0.01}     \\
            \midrule
            \midrule
            Xception                        & 99.94          & 94.47          & 87.17          \\
            +Ours, \textit{w/o\ MEB$|$FAM}  & 99.95          & 97.21          & 92.62          \\
            +Ours, \textit{w/\ \, MEB}      & 99.95          & 97.40          & 93.13          \\
            +Ours, \textit{w/\ \, FAM}      & 99.96          & 98.06          & 94.83          \\
            +Ours, \textit{w/\ \, FAM\&MEB} & \textbf{99.97} & \textbf{98.35} & \textbf{95.50} \\
            \bottomrule
        \end{tabular}
    \end{center}
    \vspace{-0.1cm}
    \caption{
        Ablation for the effect of different settings in RFM on DFFD.
        \textbf{MEB}: Multiple Erasing Blocks, \textbf{FAM}: Forgery Attention Map.}
    \label{tab:ablation}
    \vspace{-0.1cm}
\end{table}

\begin{table}[t]
    \small
    \begin{center}
        \begin{tabular}{l|l|c|c|c}
            \toprule
            Iter.                                 & Method                  & AUC            & \tdr{0.1}      & \tdr{0.01}     \\
            \midrule
            \midrule
            \multirow{2}{*}{\shortstack[l]{250k}} & +Ours, \textit{w/ PSFE} & 99.94          & \textbf{96.91} & \textbf{92.84} \\
                                                  & +Ours, \textit{w/ SFE}  & \textbf{99.95} & 96.17          & 92.78          \\
            \midrule
            \multirow{2}{*}{\shortstack[l]{350k}} & +Ours, \textit{w/ PSFE} & 99.97          & 98.25          & 95.34          \\
                                                  & +Ours, \textit{w/ SFE}  & 99.97          & \textbf{98.35} & \textbf{95.50} \\
            \bottomrule
        \end{tabular}
    \end{center}
    \vspace{-0.1cm}
    \caption{
        Comparison of SFE with PSFE on DFFD.
        We separately compared their performance under 250k and 350k training iterations.
    }
    \label{tab:ablation1}
    \vspace{-0.3cm}
\end{table}

\begin{table*}[t]
    \small
    \begin{center}
        \begin{tabular}{l|p{1.0cm}<{\centering}cc|p{1.0cm}<{\centering}cc|p{1.0cm}<{\centering}cc}
            \toprule
            \multirow{2}{*}[-0.8ex]{Method}                           &
            \multicolumn{3}{c|}{Celeb-DF}                             &
            \multicolumn{3}{c|}{DFFD (Group A)}                       &
            \multicolumn{3}{c}{DFFD (Group B)}
            \\
            \cmidrule(lr){2-4} \cmidrule(lr){5-7} \cmidrule(lr){8-10} & AUC            & \tdr{0.1}      & \tdr{0.01}     & AUC            & \tdr{0.1}      & \tdr{0.01}     & AUC            & \tdr{0.1}      & \tdr{0.01}     \\
            \midrule
            \midrule
            Xception                                                  & 99.85          & 89.11          & 84.22          & 99.94          & 97.67          & 94.57          & 99.92          & 92.87          & 83.46          \\
            +AE~\cite{wei2017object}                                  & 99.84          & 84.05          & 76.63          & 99.94          & 97.98          & 93.64          & 99.92          & 92.97          & 81.73          \\
            +RE~\cite{zhong2020random}                                & 99.89          & 88.11          & 85.20          & 99.95          & 98.35          & 95.08          & 99.96          & 96.53          & 91.89          \\
            +Ours (RFM)                                               & \textbf{99.94} & \textbf{93.88} & \textbf{87.08} & \textbf{99.97} & \textbf{99.53} & \textbf{98.91} & \textbf{99.96} & \textbf{97.76} & \textbf{93.80} \\

            \midrule
            \midrule
            Patch~\cite{chai2020makes}                                & 99.96          & 91.83          & 86.16          & 99.94          & 99.85          & 99.23          & 99.96          & \textbf{99.58} & 98.75          \\
            +Ours (RFM)                                               & \textbf{99.97} & \textbf{93.44} & \textbf{89.58} & \textbf{99.95} & \textbf{99.87} & \textbf{99.68} & \textbf{99.97} & 99.56          & \textbf{98.87} \\
            \bottomrule
        \end{tabular}
    \end{center}
    \vspace{-0.1cm}
    \caption{ Comparison of RFM with well-known erasing methods and state of the art on DFFD and Celeb-DF.
        We use Xception as backbone in the first cell and use Patch as backbone in the second cell.
    }
    \label{tab:comparison2}
    \vspace{-0.1cm}
\end{table*}


\begin{table}[t]
    \footnotesize
    \begin{center}
        \begin{tabular}{l|c|c|c|c|c}
            \toprule
            Method   & $Size$ & $p$ & AUC            & \tdr{0.1}      & \tdr{0.01}     \\
            \midrule
            \midrule 
            Xception & -      & -   & 96.87          & 40.19          & 25.14          \\ 
            +Ours    & 30     & 0.5 & \textbf{97.89} & \textbf{48.64} & \textbf{35.89} \\ 
            +Ours    & 30     & 1.0 & 97.32          & 47.23          & 29.38          \\ 
            +Ours    & 120    & 0.5 & 96.79          & 38.04          & 29.23          \\ 
            +Ours    & 120    & 1.0 & 96.41          & 34.96          & 26.48          \\ 
            \midrule
            \midrule 
            Xception & -      & -   & 96.71          & 41.94          & 34.07          \\ 
            +Ours    & 30     & 0.5 & 96.99          & 44.02          & 34.42          \\ 
            +Ours    & 30     & 1.0 & 96.93          & 43.57          & 33.75          \\ 
            +Ours    & 120    & 0.5 & \textbf{97.95} & \textbf{50.39} & 40.06          \\ 
            +Ours    & 120    & 1.0 & 97.87          & 50.17          & \textbf{40.86} \\ 
            \midrule
            \midrule 
            Xception & -      & -   & 99.36          & 70.34          & 59.10          \\
            +Ours    & 30     & 1.0 & 99.38          & 70.95          & 59.40          \\
            +Ours    & 120    & 1.0 & \textbf{99.53} & \textbf{76.26} & \textbf{62.12} \\
            \bottomrule
        \end{tabular}
    \end{center}
    \vspace{-0.1cm}
    \caption{
        Detection performance under different hyper-parameters on Celeb-DF.
        Results in the three cells are from the detectors trained on \textbf{Celeb-DF-subsetA}, \textbf{Celeb-DF-subsetB} and \textbf{Celeb-DF-subsetA\&B}, respectively.
        $p$: Erasing probability, $Size$: Max erase size $H_{max} \& W_{max}$.}
    \label{tab:ablation2}
    \vspace{0.cm}
\end{table}

\textbf{Comparison with Progressive Suspicious Forgeries Erasing.}
We also try to design an iterative erasing procedure to replace SFE.
The new method progressively occludes the most sensitive region of face rather than using multiple erasing blocks to occlude face at a time.
Concretely, we erase the Top-1 sensitive region of face under the guidance of FAM, and then re-generate FAM of the erased face for next erasing.
The two steps work in an alternative manner until the facial image has been erased $N$ times.
We call this method \textit{Progressive Suspicious Forgeries Erasing} (PSFE).
To evaluate the effectiveness of PSFE, we separately compare the performance of detectors trained under different iterations.
Although PSFE consumes three times the time of SFE in processing, the results in Table \ref{tab:ablation1} show that PSFE does not achieve significant performance gains.

\begin{figure}[t]
    \begin{center}
        \includegraphics[width=0.95\linewidth]
        {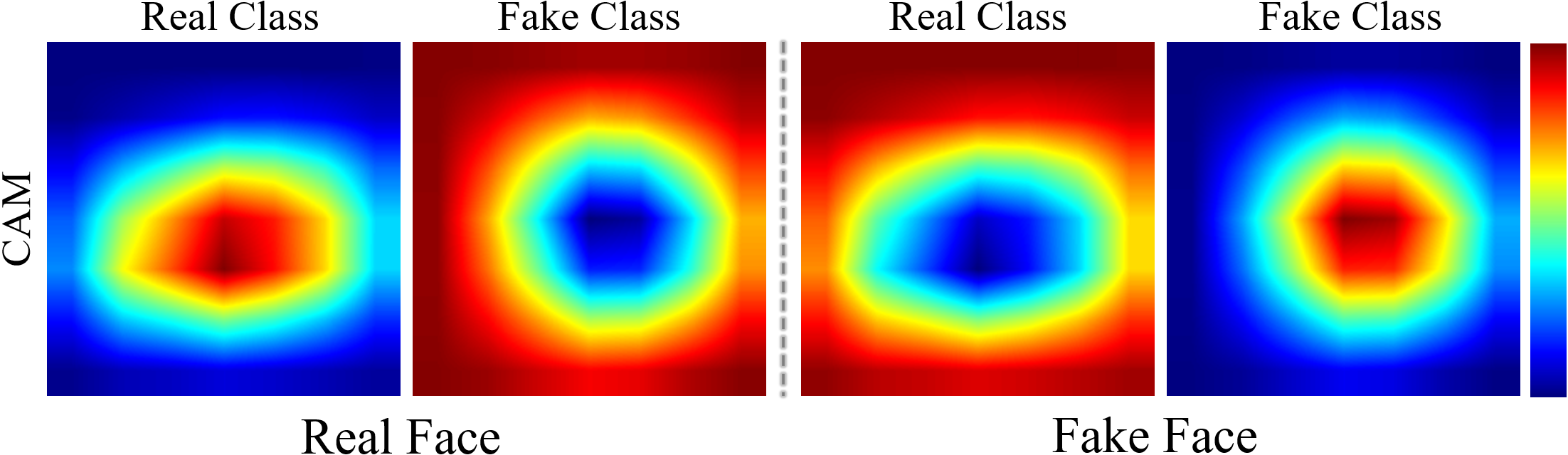}
    \end{center}
    \vspace{-0.2cm}
    \caption{
        Average Class Activation Mapping (CAM)~\cite{zhou2016learning} on 512 facial images in Celeb-DF.
        The first and second colomns represent the average CAM on real and fake faces, respectively.
    }
    \label{fig:camcmp}
    \vspace{-0.2cm}
\end{figure}

\textbf{The impact of hyper-parameters.}
To investigate the optimal hyper-parameters under different compositions of training data, we extract two sub-datasets from the training set in Celeb-DF.
The fake faces in Celeb-DF can be represented as replacing the source face with target identity.
By limiting the count of source faces and target identities in training set to 8, we can construct two sub-datasets \textbf{Celeb-DF-subsetA} and \textbf{Celeb-DF-subsetB}, respectively.
Basing on these datasets, we separately train a series of detectors with different settings on Max erase size $H_{max} \& W_{max}$ and Erasing probability $p$.
All the detectors are trained about 120k iterations, and the results are shown in Table~\ref{tab:ablation2}.
It is obvious that both the composition of training set and the setting of hyper-parameters have significant impact on RFM, and $H_{max} \& W_{max}$ is more decisive on detection performance than $p$.
Moreover, large $H_{max} \& W_{max}$ and $p$ help detector improve its detection performance when training on Celeb-DF-subsetA\&B or Celeb-DF-subsetB, while the large hyper-parameters lead to performance degradation when training on Celeb-DF-subsetA.
Unlike training with larger parameters, training with smaller parameters helps to improve performance in most cases, which means that the parameters can be set from small to large until the detector reaches the optimal performance.
Anyway, in order to achieve the optimal effect, it is essential to set proper parameters during training.

\begin{table}[t]
    \small
    \begin{center}
        \begin{tabular}{l|c|c|c}
            \toprule
            Method                            & AUC            & \tdr{0.1}      & \tdr{0.01}     \\
            \midrule
            \midrule
            Xception~\cite{dang2020detection} & 99.61          & 85.26          & 77.42          \\
            +Reg.\cite{dang2020detection}     & 99.64          & 90.78          & 83.83          \\
            \midrule
            \midrule
            Xception                          & 99.87          & 85.55          & 77.92          \\
            +Ours (RFM)                       & \textbf{99.94} & \textbf{96.94} & \textbf{90.44} \\
            \bottomrule
        \end{tabular}
    \end{center}
    \vspace{-0.1cm}
    \caption{ Comparison of RFM with state of the art on DFFD.
        Results shown in the first cell are from~\cite{dang2020detection}, and results in the second cell represent the detectors trained with the same iterations as~\cite{dang2020detection} on DFFD without DFL.}
    \label{tab:comparison1}
    \vspace{-0.2cm}
\end{table}

\begin{table*}[t]
    \footnotesize
    \begin{center}
        \begin{tabular}{l|l|ccc|ccc|ccc}
            \toprule
            \multirow{2}{*}[-0.8ex]{Part}
             & \multirow{2}{*}[-0.8ex]{Method}
             & \multicolumn{3}{c}{Celeb-DF}
             & \multicolumn{3}{c|}{DFFD (Group A)}
             & \multicolumn{3}{c}{DFFD (Group B)}
            \\
            \cmidrule(lr){3-5} \cmidrule(lr){6-8} \cmidrule(lr){9-11}
             &                                     & AUC            & \tdr{0.1}      & \tdr{0.01}     & AUC            & \tdr{0.1}      & \tdr{0.01}     & AUC            & \tdr{0.1}      & \tdr{0.01}      \\
            \midrule
            \midrule
            \multirow{2}{*}{\shortstack[l]{Eyes                                                                                                                                                              \\    Real}}
             & Xception                            & 69.99          & 00.83          & 00.39          & 93.41          & 33.62          & 20.04          & 97.58          & 44.88          & 22.79           \\
             & +Ours (RFM)                         & \textbf{93.43} & \textbf{20.47} & \textbf{13.86} & \textbf{99.72} & \textbf{78.19} & \textbf{52.63} & \textbf{99.30} & \textbf{61.12} & \textbf{29.67}  \\
            \midrule
            \multirow{2}{*}{\shortstack[l]{Nose                                                                                                                                                              \\    Real}}
             & Xception                            & 82.77          & 02.92          & 01.33          & 95.50          & 59.99          & 45.59          & 98.51          & 76.01          & 59.50           \\
             & +Ours (RFM)                         & \textbf{98.88} & \textbf{48.33} & \textbf{35.19} & \textbf{99.95} & \textbf{98.20} & \textbf{95.90} & \textbf{99.92} & \textbf{93.45} & \textbf{85.58}  \\
            \midrule
            \multirow{2}{*}{\shortstack[l]{Mouth                                                                                                                                                             \\   Real}}
             & Xception                            & 98.30          & 46.18          & 32.36          & 98.76          & 67.93          & 49.94          & 99.25          & 66.61          & 48.44           \\
             & +Ours (RFM)                         & \textbf{98.50} & \textbf{46.30} & \textbf{34.83} & \textbf{99.95} & \textbf{97.77} & \textbf{94.39} & \textbf{99.84} & \textbf{86.37} & \textbf {75.73} \\
            \midrule
            \multirow{2}{*}{\shortstack[l]{Skin                                                                                                                                                              \\    Real}}
             & Xception                            & 85.33          & 08.20          & 04.53          & 92.73          & 16.81          & 6.779          & 94.95          & 47.38          & 29.24           \\
             & +Ours (RFM)                         & \textbf{97.87} & \textbf{45.44} & \textbf{34.36} & \textbf{93.79} & \textbf{35.80} & \textbf{28.97} & \textbf{96.80} & \textbf{61.06} & \textbf{48.52}  \\
            \midrule
            \midrule
            \multirow{2}{*}{\shortstack[l]{Eyes                                                                                                                                                              \\    Real}}
             & Patch                               & 97.06          & 02.25          & 01.28          & 99.91          & 87.86          & 12.70          & 99.87          & 60.64          & 06.92           \\
             & +Ours (RFM)                         & \textbf{97.82} & \textbf{06.56} & \textbf{04.69} & \textbf{99.93} & \textbf{97.21} & \textbf{20.56} & \textbf{99.90} & \textbf{81.89} & \textbf{28.40}  \\
            \midrule
            \multirow{2}{*}{\shortstack[l]{Nose                                                                                                                                                              \\    Real}}
             & Patch                               & 98.76          & 21.75          & 13.98          & 99.97          & 99.64          & 89.83          & 99.96          & 97.64          & 68.10           \\
             & +Ours (RFM)                         & \textbf{99.52} & \textbf{56.17} & \textbf{47.21} & \textbf{99.97} & \textbf{99.84} & \textbf{97.91} & \textbf{99.97} & \textbf{97.68} & \textbf{84.17}  \\
            \midrule
            \multirow{2}{*}{\shortstack[l]{Mouth                                                                                                                                                             \\   Real}}
             & Patch                               & \textbf{99.32} & 12.68          & 06.46          & 99.96          & 99.81          & 87.86          & 99.96          & 99.16          & 65.08           \\
             & +Ours (RFM)                         & \textbf{99.32} & \textbf{29.84} & \textbf{23.03} & \textbf{99.97} & \textbf{99.86} & \textbf{95.28} & \textbf{99.97} & \textbf{99.22} & \textbf{83.65}  \\
            \midrule
            \multirow{2}{*}{\shortstack[l]{Skin                                                                                                                                                              \\    Real}}
             & Patch                               & 98.98          & 29.85          & 15.26          & 98.88          & 59.18          & 25.06          & \textbf{99.87} & 80.81          & 20.01           \\
             & +Ours (RFM)                         & \textbf{99.55} & \textbf{47.30} & \textbf{32.70} & \textbf{99.66} & \textbf{77.52} & \textbf{39.47} & 99.73          & \textbf{80.89} & \textbf{39.77}  \\
            \bottomrule
        \end{tabular}
    \end{center}
    \vspace{-0.1cm}
    \caption{ Comparison of RFM with baseline methods on the facial images which only have few technical forgeries.}
    \label{tab:withoutregion}
    \vspace{-0.2cm}
\end{table*}

\subsection{Experiment on DFFD and Celeb-DF}\label{sec:detectionresults}
\textbf{Comparison with well-known erasing methods.}
We separately conduct experiments on DFFD and Celeb-DF to compare RFM with well-known erasing methods such as Adversarial Erasing (AE)~\cite{wei2017object} and Random Erasing (RE)~\cite{zhong2020random}.
Random integers are used to compose erasing blocks for all three methods.
And the hyper-parameters in both AE and RE are set as the original settings.

The results in Table~\ref{tab:comparison2} shows that RFM outperforms baseline and other erasing methods by a large margin.
Meanwhile, it is counter-intuitive that the usage of AE leads to performance degradation.
To further figure out why AE does not work on fake face detection, we generate the average CAM on 512 real and fake faces separately.
As shown in Figure~\ref{fig:camcmp}, we find that the CAM which AE used only contains high-dimensional information and loses the representation ability for forgery region, providing insufficient guidance for erasing.

\textbf{Comparison with state of the art.}
In order to further evaluate the effectiveness of RFM, we make comparisons with state of the art on DFFD and Celeb-DF separately.
As shown in Table~\ref{tab:comparison2}, a vanilla Xception with RFM can achieve competitive performance with Patch~\cite{chai2020makes}.
Moreover, by using RFM generated images for training, Patch achieves virtually the best performance than any other methods.
Obviously, it is also a strong proof that RFM can be easily integrated with various models to improve fake face detection performance.

\begin{figure}[t]
    \begin{center}
        \includegraphics[width=0.99\linewidth]
        {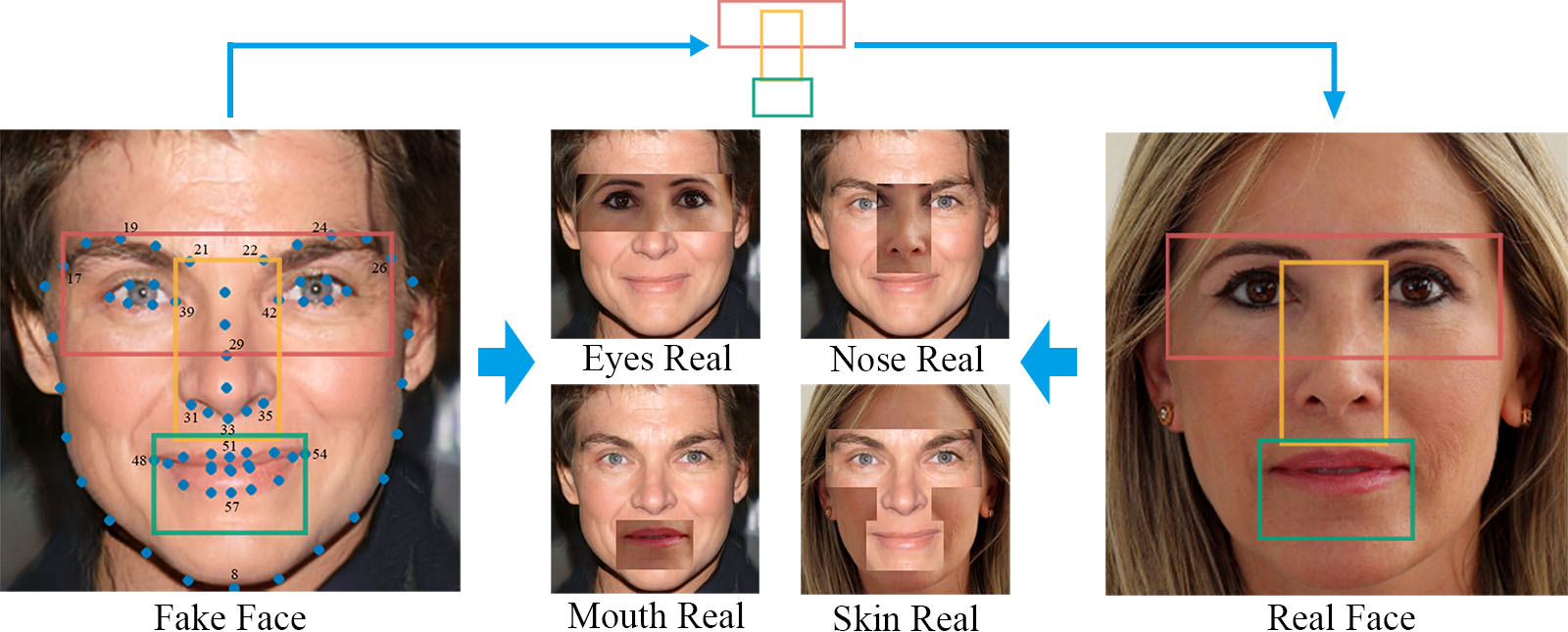}
    \end{center}
    \vspace{-0.2cm}
    \caption{
        The generation procedure of less-forgery fake faces.
        The image with suffix ``Real'' means that the region in fake face have been replaced with the corresponding pixels from a random real face.
    }
    \label{fig:faceseg}
    \vspace{-0.4cm}
\end{figure}

\begin{figure*}[t]
    \begin{center}
        \includegraphics[width=0.98\linewidth]{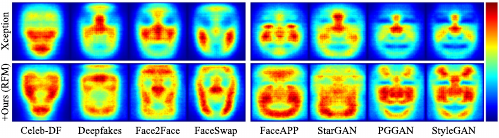}
    \end{center}
    \vspace{-0.3cm}
    \caption{
        Average FAMs separately generated under the detectors trained with different methods.
        The manipulation techniques in the left and right colomn are consist of two-stage and one-stage techniques respectively.
    }
    \label{fig:mapcomparison}
    \vspace{-0.1cm}
\end{figure*}

\begin{figure}[t]
    \begin{center}
        \includegraphics[width=1.\linewidth]
        {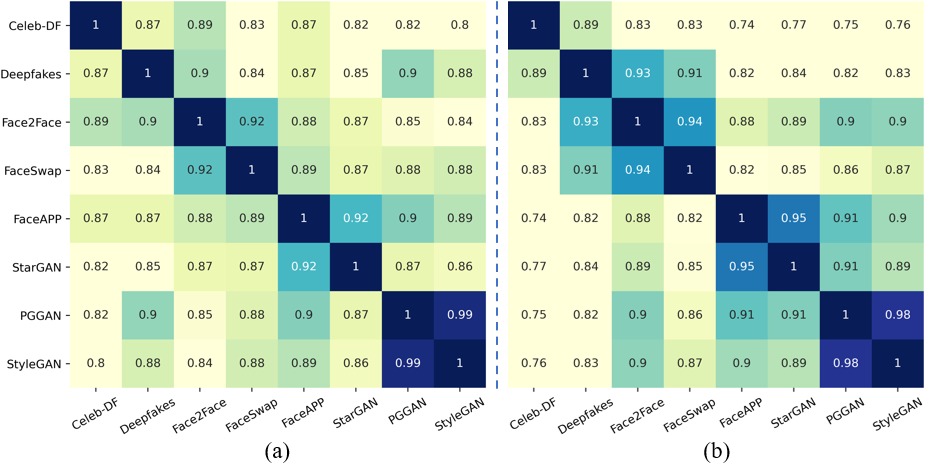}
    \end{center}
    \vspace{-0.3cm}
    \caption{
        The correlation matrixes calculate the normalized cosine similarity of average FAMs between each pair of manipulation techniques.
        The average FAMs are generated under the detectors trained (a) with or (b) without RFM respectively.
    }
    \label{fig:hotmap}
    \vspace{-0.3cm}
\end{figure}

Additionally, we also compare RFM with~\cite{dang2020detection}.
Since the DFFD we collected lacks fake faces from Deep Face Lab (DFL)~\cite{deepfacelab}, we implement the same model selection strategy as that in~\cite{dang2020detection} to conduct a roughly fair comparison.
The whole process can be divided into two steps:
Firstly, following the training iteration in~\cite{dang2020detection}, we trained an Xception-based detector on the incomplete DFFD until the detector achieves the same \tdr{0.1} and \tdr{0.01} as~\cite{dang2020detection}.
Then, we utilize RFM to train another detector under the same iteration.
The results in Table~\ref{tab:comparison1} indicate that RFM yields improvements of 6.16\% \tdr{0.1} and 6.61\% \tdr{0.01} when compared with results in~\cite{dang2020detection}.

\textbf{Robustness on less-forgery fake faces.}
To explore how RFM affects the detection performance on the fake faces with few technical forgeries, we propose to leverage semantic-based segmentation to generate less-forgery fake faces for testing.
Concretely, we firstly obtain the locations of 68 facial landmarks by utilizing the facial landmarks extractor in dlib~\cite{king2009dlib}.
Then, the facial landmarks are used to divide fake face into four parts: eyes, nose, mouth and facial skin.
Next, less-forgery fake faces are generated by separately replacing each region in fake face with the corresponding pixels of a real face.
Finally, we construct four datasets that contain the fake faces with few technical forgeries on eyes, nose, mouth, and facial skin region respectively.
The detailed process is shown in Figure~\ref{fig:faceseg}.

We separately evaluate the detectors trained with or without RFM on these datasets.
The results in Table \ref{tab:withoutregion} demonstrate that the detectors without RFM encounter severe performance degradation when facing less-forgery faces, while REF effectively helps detector maintain performance on less-forgery faces.

\subsection{Representative Forgery Visualization}\label{sec:visualization}

\textbf{Visualization on several images.}
In order to exhibit the representative forgery region discovered by RFM, we generate average FAM on 512 fake facial images for each face manipulation technique.
As illustarted in Figure~\ref{fig:mapcomparison}, the detector trained with RFM pays attention to a more comprehensive and representative region of interest, where the facial boundary of faces generated by two-stage techniques and the entire skin of faces generated by one-stage techniques.


Furthermore, under the help of RFM, the average FAMs of fake faces generated by similar techniques tend to be similar to each other (as the correlation matrixes shown in Figure~\ref{fig:hotmap}), which reflects that RFM produces a clustering effect on the fake faces of similar techniques.
Therefore, our method can be further utilized to explore the technical procedure of a black-box manipulation technique.
To achieve this goal, we firstly train a detector with RFM on the fake faces generated by both known and unknown techniques.
Then, we generate the average FAM for each technique and calculate the normalized cosine similarity of FAMs between each pair of known and unknown techniques.
After that, the technical procedure of a black-box technique can be determined according to its correlation with other known methods.
As shown in the matrix, it can be inferred that FaceAPP is belonging to one-stage manipulation technique.

\begin{figure}[t]
    \begin{center}
        \includegraphics[width=1.\linewidth]
        {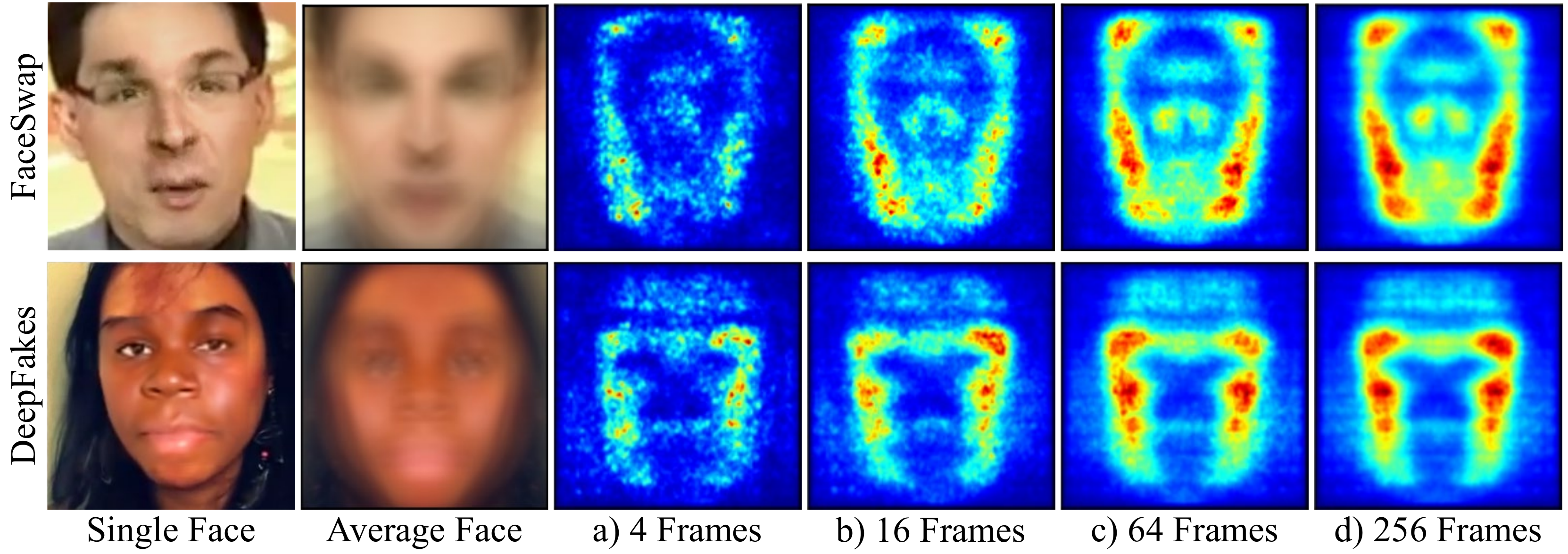}
    \end{center}
    \vspace{-0.2cm}
    \caption{
        Comparison of average FAMs calculated on a single fake video with different frames.
        `Single face' shows the fake face of single frame and `Average face' shows the average fake face of multiple frames.
        The columns a) to d) shows the average FAMs calculated on 4 to 256 frames, respectively.
    }
    \label{fig:videoFAM}
    \vspace{-0.2cm}
\end{figure}

\textbf{Visualization on a single video.}
Actually, fake faces generated by two-stage techniques mainly appear in the form of video.
To further investigate how the number of video frames of single fake video influences on the visualization effectiveness of our method, we exhibit the average FAMs generated based on different frames of a single video.
As shown in Figure~\ref{fig:videoFAM}, average FAM generated on 4 frames is enough to show the representative forgery with discriminative contour, and the complete contour and inner of representative forgery appear gradually when the number of frames raises from 4 to 256.
In conclusion, RFM performs well on representative forgery mining.

\section{Conclusion}
In this work, we provide insight into fake face detection that detection performance can be effectively improved by refining training data.
Concretely, we propose a novel attention-based data augmentation method to guide detector explore representative forgery from the previously ignored facial region.
In addition, the visualization result shows that our method can separately discover the corresponding representative forgery of different manipulation techniques without the need of well-designed supervision.
With our method, a vanilla CNN-based detector can achieve state-of-the-art performance on the well-known fake face datasets DFFD and Celeb-DF.

\noindent
\textbf{Acknowledgment.}
This work was supported by National Key R\&D Program of China (2019YFB1406504).

    {\small
        \bibliographystyle{ieee_fullname}
        \bibliography{cvpr}

\begin{thebibliography}{10}\itemsep=-1pt

\bibitem{deepfacelab}
Deepfacelab.
\newblock \url{https://github.com/iperov/DeepFaceLab}.
\newblock Accessed: 2019-09-04.

\bibitem{deepfakes}
Deepfakes github.
\newblock \url{https://github.com/deepfakes/faceswap}.
\newblock Accessed: 2018-10-29.

\bibitem{faceapp}
Faceapp.
\newblock \url{https://faceapp.com/app}.
\newblock Accessed: 2019-09-04.

\bibitem{faceswap}
Faceswap.
\newblock \url{https://github.com/MarekKowalski/FaceSwap}.
\newblock Accessed: 2018-10-29.

\bibitem{thispersondoesnotexist}
This person does not exist.
\newblock \url{https://thispersondoesnotexist.com}.
\newblock Accessed: 2019-2-15.

\bibitem{chai2020makes}
Lucy Chai, David Bau, Ser-Nam Lim, and Phillip Isola.
\newblock What makes fake images detectable? understanding properties that
  generalize.
\newblock In {\em European Conference on Computer Vision}, 2020.

\bibitem{chattopadhay2018grad}
Aditya Chattopadhay, Anirban Sarkar, Prantik Howlader, and Vineeth~N
  Balasubramanian.
\newblock Grad-cam++: Generalized gradient-based visual explanations for deep
  convolutional networks.
\newblock In {\em 2018 IEEE Winter Conference on Applications of Computer
  Vision (WACV)}, pages 839--847. IEEE, 2018.

\bibitem{chen2020manipulated}
Zehao Chen and Hua Yang.
\newblock Manipulated face detector: Joint spatial and frequency domain
  attention network.
\newblock {\em arXiv preprint arXiv:2005.02958}, 2020.

\bibitem{choi2018stargan}
Yunjey Choi, Minje Choi, Munyoung Kim, Jung-Woo Ha, Sunghun Kim, and Jaegul
  Choo.
\newblock Stargan: Unified generative adversarial networks for multi-domain
  image-to-image translation.
\newblock In {\em Proceedings of the IEEE conference on computer vision and
  pattern recognition}, pages 8789--8797, 2018.

\bibitem{chollet2017xception}
Fran{\c{c}}ois Chollet.
\newblock Xception: Deep learning with depthwise separable convolutions.
\newblock In {\em Proceedings of the IEEE conference on computer vision and
  pattern recognition}, pages 1251--1258, 2017.

\bibitem{dang2020detection}
Hao Dang, Feng Liu, Joel Stehouwer, Xiaoming Liu, and Anil~K Jain.
\newblock On the detection of digital face manipulation.
\newblock In {\em Proceedings of the IEEE/CVF Conference on Computer Vision and
  Pattern Recognition}, pages 5781--5790, 2020.

\bibitem{devries2017improved}
Terrance DeVries and Graham~W Taylor.
\newblock Improved regularization of convolutional neural networks with cutout.
\newblock {\em arXiv preprint arXiv:1708.04552}, 2017.

\bibitem{huang2020fakelocator}
Yihao Huang, Felix Juefei-Xu, Run Wang, Xiaofei Xie, Lei Ma, Jianwen Li, Weikai
  Miao, Yang Liu, and Geguang Pu.
\newblock Fakelocator: Robust localization of gan-based face manipulations via
  semantic segmentation networks with bells and whistles.
\newblock {\em arXiv preprint arXiv:2001.09598}, 2020.

\bibitem{ishida2020we}
Takashi Ishida, Ikko Yamane, Tomoya Sakai, Gang Niu, and Masashi Sugiyama.
\newblock Do we need zero training loss after achieving zero training error?
\newblock In {\em International Conference on Machine Learning}, pages
  4604--4614. PMLR, 2020.

\bibitem{karras2018progressive}
Tero Karras, Timo Aila, Samuli Laine, and Jaakko Lehtinen.
\newblock Progressive growing of gans for improved quality, stability, and
  variation.
\newblock In {\em International Conference on Learning Representations}, 2018.

\bibitem{karras2019style}
Tero Karras, Samuli Laine, and Timo Aila.
\newblock A style-based generator architecture for generative adversarial
  networks.
\newblock In {\em Proceedings of the IEEE conference on computer vision and
  pattern recognition}, pages 4401--4410, 2019.

\bibitem{karras2020analyzing}
Tero Karras, Samuli Laine, Miika Aittala, Janne Hellsten, Jaakko Lehtinen, and
  Timo Aila.
\newblock Analyzing and improving the image quality of stylegan.
\newblock In {\em Proceedings of the IEEE/CVF Conference on Computer Vision and
  Pattern Recognition}, pages 8110--8119, 2020.

\bibitem{king2009dlib}
Davis~E King.
\newblock Dlib-ml: A machine learning toolkit.
\newblock {\em The Journal of Machine Learning Research}, 10:1755--1758, 2009.

\bibitem{kingma2015adam}
Diederik~P Kingma and Jimmy Ba.
\newblock Adam: A method for stochastic optimization.
\newblock In {\em International Conference on Learning Representations}, 2015.

\bibitem{krizhevsky2012imagenet}
Alex Krizhevsky, Ilya Sutskever, and Geoffrey~E Hinton.
\newblock Imagenet classification with deep convolutional neural networks.
\newblock In {\em Advances in neural information processing systems}, pages
  1097--1105, 2012.

\bibitem{li2020advancing}
Lingzhi Li, Jianmin Bao, Hao Yang, Dong Chen, and Fang Wen.
\newblock Advancing high fidelity identity swapping for forgery detection.
\newblock In {\em Proceedings of the IEEE/CVF Conference on Computer Vision and
  Pattern Recognition}, pages 5074--5083, 2020.

\bibitem{li2020face}
Lingzhi Li, Jianmin Bao, Ting Zhang, Hao Yang, Dong Chen, Fang Wen, and Baining
  Guo.
\newblock Face x-ray for more general face forgery detection.
\newblock In {\em Proceedings of the IEEE/CVF Conference on Computer Vision and
  Pattern Recognition}, pages 5001--5010, 2020.

\bibitem{li2020celeb}
Yuezun Li, Xin Yang, Pu Sun, Honggang Qi, and Siwei Lyu.
\newblock Celeb-df: A large-scale challenging dataset for deepfake forensics.
\newblock In {\em Proceedings of the IEEE/CVF Conference on Computer Vision and
  Pattern Recognition}, pages 3207--3216, 2020.

\bibitem{liu2020global}
Zhengzhe Liu, Xiaojuan Qi, and Philip~HS Torr.
\newblock Global texture enhancement for fake face detection in the wild.
\newblock In {\em Proceedings of the IEEE/CVF Conference on Computer Vision and
  Pattern Recognition}, pages 8060--8069, 2020.

\bibitem{nataraj2019detecting}
Lakshmanan Nataraj, Tajuddin~Manhar Mohammed, BS Manjunath, Shivkumar
  Chandrasekaran, Arjuna Flenner, Jawadul~H Bappy, and Amit~K Roy-Chowdhury.
\newblock Detecting gan generated fake images using co-occurrence matrices.
\newblock {\em Electronic Imaging}, 2019(5):532--1, 2019.

\bibitem{roessler2018faceforensics}
Andreas R\"ossler, Davide Cozzolino, Luisa Verdoliva, Christian Riess, Justus
  Thies, and Matthias Nie{\ss}ner.
\newblock Face{F}orensics: A large-scale video dataset for forgery detection in
  human faces.
\newblock {\em arXiv}, 2018.

\bibitem{roessler2019faceforensicspp}
Andreas R\"ossler, Davide Cozzolino, Luisa Verdoliva, Christian Riess, Justus
  Thies, and Matthias Nie{\ss}ner.
\newblock Face{F}orensics++: Learning to detect manipulated facial images.
\newblock In {\em International Conference on Computer Vision (ICCV)}, 2019.

\bibitem{rossler2019faceforensics}
Andreas Rossler, Davide Cozzolino, Luisa Verdoliva, Christian Riess, Justus
  Thies, and Matthias Nie{\ss}ner.
\newblock Faceforensics++: Learning to detect manipulated facial images.
\newblock In {\em Proceedings of the IEEE International Conference on Computer
  Vision}, pages 1--11, 2019.

\bibitem{selvaraju2017grad}
Ramprasaath~R Selvaraju, Michael Cogswell, Abhishek Das, Ramakrishna Vedantam,
  Devi Parikh, and Dhruv Batra.
\newblock Grad-cam: Visual explanations from deep networks via gradient-based
  localization.
\newblock In {\em Proceedings of the IEEE international conference on computer
  vision}, pages 618--626, 2017.

\bibitem{shu2017fake}
Kai Shu, Amy Sliva, Suhang Wang, Jiliang Tang, and Huan Liu.
\newblock Fake news detection on social media: A data mining perspective.
\newblock {\em ACM SIGKDD explorations newsletter}, 19(1):22--36, 2017.

\bibitem{simonyan2014very}
Karen Simonyan and Andrew Zisserman.
\newblock Very deep convolutional networks for large-scale image recognition.
\newblock In {\em International Conference on Learning Representations}, 2015.

\bibitem{srivastava2014dropout}
Nitish Srivastava, Geoffrey Hinton, Alex Krizhevsky, Ilya Sutskever, and Ruslan
  Salakhutdinov.
\newblock Dropout: a simple way to prevent neural networks from overfitting.
\newblock {\em The journal of machine learning research}, 15(1):1929--1958,
  2014.

\bibitem{suwajanakorn2017synthesizing}
Supasorn Suwajanakorn, Steven~M Seitz, and Ira Kemelmacher-Shlizerman.
\newblock Synthesizing obama: learning lip sync from audio.
\newblock {\em ACM Transactions on Graphics (TOG)}, 36(4):1--13, 2017.

\bibitem{thies2016face2face}
Justus Thies, Michael Zollhofer, Marc Stamminger, Christian Theobalt, and
  Matthias Nie{\ss}ner.
\newblock Face2face: Real-time face capture and reenactment of rgb videos.
\newblock In {\em Proceedings of the IEEE conference on computer vision and
  pattern recognition}, pages 2387--2395, 2016.

\bibitem{tolosana2020deepfakes}
Ruben Tolosana, Ruben Vera-Rodriguez, Julian Fierrez, Aythami Morales, and
  Javier Ortega-Garcia.
\newblock Deepfakes and beyond: A survey of face manipulation and fake
  detection.
\newblock {\em arXiv preprint arXiv:2001.00179}, 2020.

\bibitem{wang2020cnn}
Sheng-Yu Wang, Oliver Wang, Richard Zhang, Andrew Owens, and Alexei~A Efros.
\newblock Cnn-generated images are surprisingly easy to spot... for now.
\newblock In {\em Proceedings of the IEEE Conference on Computer Vision and
  Pattern Recognition}, volume~7, 2020.

\bibitem{wei2017object}
Yunchao Wei, Jiashi Feng, Xiaodan Liang, Ming-Ming Cheng, Yao Zhao, and
  Shuicheng Yan.
\newblock Object region mining with adversarial erasing: A simple
  classification to semantic segmentation approach.
\newblock In {\em Proceedings of the IEEE conference on computer vision and
  pattern recognition}, pages 1568--1576, 2017.

\bibitem{yu2019attributing}
Ning Yu, Larry~S Davis, and Mario Fritz.
\newblock Attributing fake images to gans: Learning and analyzing gan
  fingerprints.
\newblock In {\em Proceedings of the IEEE International Conference on Computer
  Vision}, pages 7556--7566, 2019.

\bibitem{zhang2018mixup}
Hongyi Zhang, Moustapha Cisse, Yann~N Dauphin, and David Lopez-Paz.
\newblock mixup: Beyond empirical risk minimization.
\newblock {\em International Conference on Learning Representations}, 2018.

\bibitem{zhang2019detecting}
Xu Zhang, Svebor Karaman, and Shih-Fu Chang.
\newblock Detecting and simulating artifacts in gan fake images.
\newblock In {\em 2019 IEEE International Workshop on Information Forensics and
  Security (WIFS)}, pages 1--6. IEEE, 2019.

\bibitem{zhong2020random}
Zhun Zhong, Liang Zheng, Guoliang Kang, Shaozi Li, and Yi Yang.
\newblock Random erasing data augmentation.
\newblock In {\em AAAI}, pages 13001--13008, 2020.

\bibitem{zhou2016learning}
Bolei Zhou, Aditya Khosla, Agata Lapedriza, Aude Oliva, and Antonio Torralba.
\newblock Learning deep features for discriminative localization.
\newblock In {\em Proceedings of the IEEE conference on computer vision and
  pattern recognition}, pages 2921--2929, 2016.

\end{thebibliography}
    }

\end{document}